\documentclass[10pt,twocolumn,letterpaper]{article}

\usepackage{iccv}              

\newcommand{\PAR}[1]{\vskip4pt \noindent{\bf #1~}}

\newcommand{\tablestyle}[2]{\setlength{\tabcolsep}{#1}\renewcommand{\arraystretch}{#2}\centering\footnotesize}

\definecolor{iccvblue}{rgb}{0.21,0.49,0.74}
\usepackage[pagebackref,breaklinks,colorlinks,allcolors=iccvblue]{hyperref}

\def\paperID{8364} 
\def\confName{ICCV}
\def\confYear{2025}

\title{Disentangled Clothed Avatar Generation with Layered Representation}

\author{
        Weitian Zhang$^{1}$
	~~
        Yichao Yan$^{1}$$^{\dagger}$
	~~ 
	Sijing Wu$^{1}$
	~~ 
	Manwen Liao$^{2}$ 
    ~~
	Xiaokang Yang$^{1}$ \\
        $^{1}$
        MoE Key Lab of Artificial Intelligence, AI Institute, 
        Shanghai Jiao Tong University \\
        $^{2}$The University of Hong Kong \\
{\tt\small $^{1}$\{weitianzhang, yanyichao, wusijing, xkyang\}@sjtu.edu.cn} \\
{\tt\small $^{2}$\{manwen\}@connect.hku.hk}
}

\newcommand\blfootnote[1]{
  \begingroup
    \renewcommand\thefootnote{}
    \footnote{#1}
    \addtocounter{footnote}{-1}
  \endgroup
}

\begin{document}
\twocolumn[{
\renewcommand\twocolumn[1][]{#1}%
\maketitle
\begin{center}
\includegraphics[width=\textwidth]{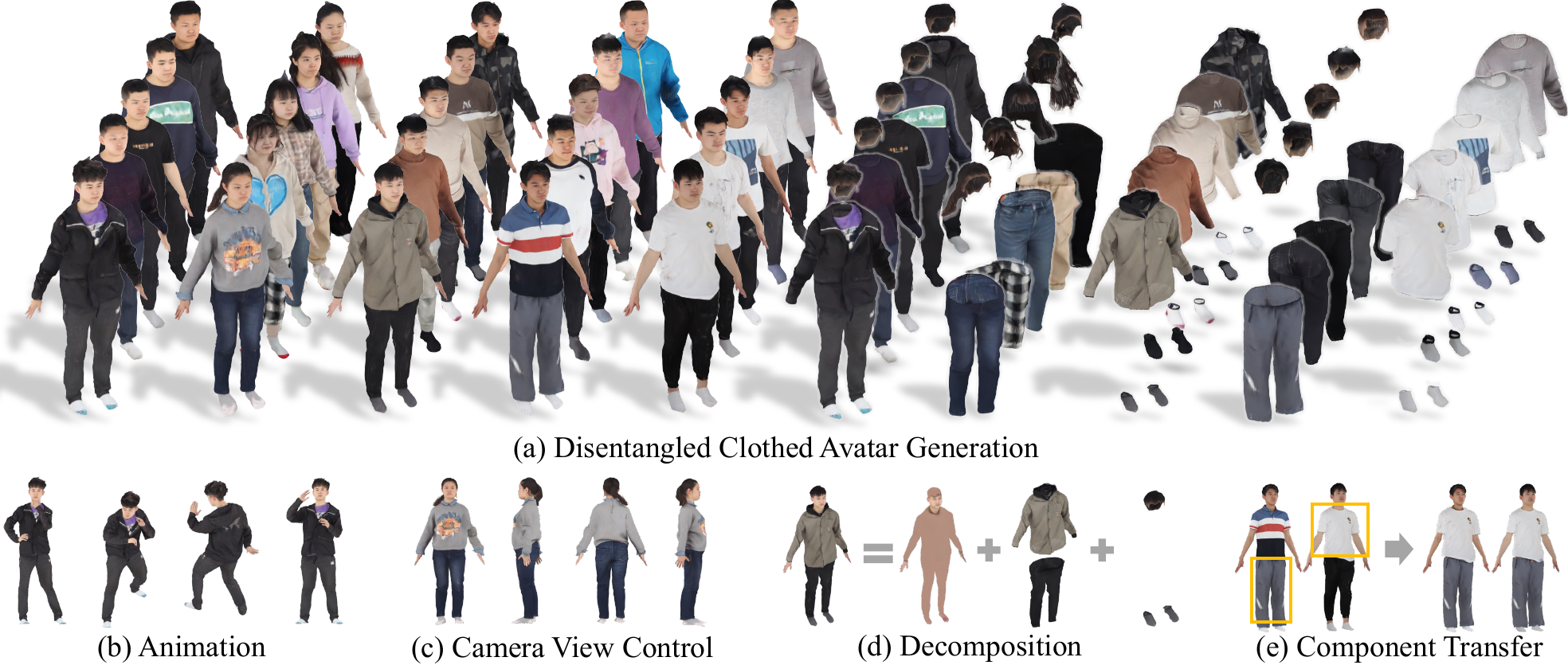}
\vspace{-0.2in}
\captionof{figure}{We propose \textbf{LayerAvatar} to efficiently generate diverse clothed avatars with components fully disentangled. The generated avatars can be animated and synthesized in novel views. They can also be decomposed into body, hair, and clothes for component transfer.}
\vspace{0in}
\label{fig:teaser}
\end{center}
}]

\blfootnote{$\dagger$ Corresponding author.}

\begin{abstract}
Clothed avatar generation has wide applications in virtual and augmented reality, filmmaking, and more. While existing methods have made progress in creating animatable digital avatars, generating avatars with disentangled components (\eg, body, hair, and clothes) has long been a challenge. In this paper, we propose LayerAvatar, a novel feed-forward diffusion-based method capable of generating high-quality component-disentangled clothed avatars in seconds. We propose a layered UV feature plane representation, where components are distributed in different layers of the Gaussian-based UV feature plane with corresponding semantic labels. This representation can be effectively learned with current feed-forward generation pipelines, facilitating component disentanglement and enhancing details of generated avatars. Based on the well-designed representation, we train a single-stage diffusion model and introduce constrain terms to mitigate the severe occlusion issue of the innermost human body layer. Extensive experiments demonstrate the superior performances of our method in generating highly detailed and disentangled clothed avatars. In addition, we explore its applications in component transfer. The project page is available at \href{https://olivia23333.github.io/LayerAvatar}{https://olivia23333.github.io/LayerAvatar}.
\end{abstract}    
\section{Introduction}
\label{sec:intro}
The creation of digital avatars has various applications~\cite{chu2020expressive,fullbody_app_2021} in virtual and augmented reality, filmmaking, and more. Traditional graphics-based pipelines require extensive effort from 3D artists to construct a single digital avatar. To reduce tedious manual labor and facilitate mass production, learning-based methods aiming at generating digital avatars automatically have been widely explored recently.

Recent learning-based methods~\cite{bergman2022gnarf, dong2023ag3d, zhang2022avatargen} mainly combine 3D representations~\cite{mildenhall2021nerf, wang2021neus, shen2021dmtet, kerbl3Dgaussians} with generation pipelines (\eg, 3D-aware GANs~\cite{GIRAFFE, piGAN2021, Chan2022eg3d} and diffusion models~\cite{rombach2021sd,peebles2023dit}) to create digital avatars. However, most of these methods often ignore the compositional nature of digital avatars and represent the human body, hair, and clothes as a whole, which limits their capabilities in digital avatar customization such as cloth transfer. 
Neural-ABC~\cite{chen2024neuralabc} and SMPLicit~\cite{corona2021smplicit} provide parametric model with disentangled clothes and human body. However, modeling texture are leaved an un-explored problem. HumanLiff~\cite{HumanLiff} proposes a layer-wise generation process that first generates clothed avatars in minimal clothes, then generates digital avatars wearing the next layer of clothing conditioned on the current layer. Nevertheless, the human body and clothes are not fully disentangled, which makes it difficult to extract the components of each layer, thus reducing the editing ability.
In addition, some methods~\cite{wang2023disentangled, dong2024tela, sun2024barbie, wang2024humancoser, gong2024laga} follow the trend of DreamFusion~\cite{poole2022dreamfusion} to achieve disentangled clothed avatar generation by learning each component of digital avatars separately through the prior knowledge of 2D diffusion models~\cite{rombach2021sd} in an optimization manner. These methods can generate clothed avatars with each component disentangled, however, they take hours to generate a single digital avatar, and the optimization time will increase linearly according to the number of components.

In this paper, we propose \textbf{LayerAvatar}, a novel feed-forward diffusion-based method that achieves (1) \textit{component disentanglement}, enabling seamless transfer of individual components such as clothes, hair, and shoes; (2) \textit{high-quality results}, with generated avatars exhibiting intricate facial details, distinct fingers, and realistic cloth wrinkles; and (3) \textit{efficiency}, requiring only seconds to generate a single avatar. 
We choose 3D Gaussians~\cite{kerbl3Dgaussians} as the underlying representation due to its high-quality rendering results for intricate details, and strong representation capability for diverse cloth types. 
However, naively representing the disentangled clothed avatar using 3D Gaussians is impractical due to its unstructured nature that is incompatible with most current feed-forward generation pipelines~\cite{Chan2022eg3d,rombach2021sd}.
Therefore, we introduce a Gaussian-based UV feature plane, in which 3D Gaussians are projected into a predefined 2D UV space shared among subjects. The attributes of each 3D Gaussian are encoded as local geometry and texture latent features, which can be obtained from the 2D feature plane via bilinear interpolation.
Furthermore, to achieve full disentanglement of avatar components (hair, shoes, upper cloth) and higher generation quality, we represent avatar components in separate layers of the UV feature plane which provides neighboring components with distinctive features from different layers to facilitate decomposition.

To generate the layered representation in a feed-forward manner, we elaborately train a single-stage diffusion model \cite{ssdnerf} from multi-view 2D images.
To fully disentangle each component and ensure plausible avatar generation results, we employ supervision both in the individual components and the entire compositional clothed avatar. 
Moreover, several prior losses are utilized to constrain the smooth surface and reasonable color of the severely occluded human body.

We evaluate LayerAvatar on multiple datasets~\cite{tao2021function4d, chen2021tightcap,ho2023custom}, demonstrating its superior performance in generating disentangled avatars. We also explore its application in component transfer. In summary, our main contributions are:
\begin{itemize}
    \item We introduce LayerAvatar, a novel feed-forward clothed avatar generation pipeline with each component disentangled, enhancing the controllability of avatar generation.
    \item We propose a layered UV feature plane representation that enhances generation quality and facilitates the disentanglement of each component.
    \item Our method achieves outstanding generation results on multiple datasets and support downstream applications such as component transfer.
\end{itemize}
\section{Related work}
\label{sec:related}

\noindent \textbf{Diffusion in 3D Generation.} 
Encouraged by the success of diffusion model~\cite{rombach2021sd} in 2D image generation area, researchers have attempted to extend it to 3D generation tasks. These works can be divided into two categories, feed-forward and optimization-based methods. Optimization-based methods~\cite{poole2022dreamfusion, chen2023fantasia3d, lin2023magic3d, wang2024prolificdreamer, qiu2024richdreamer, tang2023dreamgaussian, liang2024luciddreamer}, represented by DreamFusion~\cite{poole2022dreamfusion}, utilize SDS loss to distill prior knowledge of 2D diffusion model to supervise 3D scenes. These methods often suffer from oversaturation and Janus problems. Thus, improved SDS loss~\cite{wang2024prolificdreamer} and camera-conditional~\cite{liu2023zero123} or multi-view diffusion models~\cite{shi2023mvdream} are introduced to mitigate these problems. Moreover, optimization-based methods usually take hours to generate a single object, which hinders its application in real life. On the other hand, feed-forward methods~\cite{Liu2023MeshDiffusion, yu2023surfD, zhao2023michelangelo, zhang2024clay, wu2024direct3d} directly learn diffusion model for 3D representations, such as points~\cite{nichol2022pointe, zeng2022lion}, voxels~\cite{ren2024xcube}, meshes~\cite{Liu2023MeshDiffusion, yan2024omages64}, and implicit neural representations~\cite{muller2023diffrf, shue2023triplanediff, hong20243dtopia}. These methods can generate 3D objects in seconds. Several attempts~\cite{chen2023primdiffusion, Joint2Human, hu2024structldm, zhang2024e3gen} have been made to adapt them to the field of digital human generation. Different from most previous works, we regard digital avatars as a composition of multiple components instead of a unified whole and learn a diffusion model on the proposed layered representation. 

\noindent \textbf{Clothed Avatar Generation.}
\begin{figure*}[h]
  \centering
   \includegraphics[width=\linewidth]{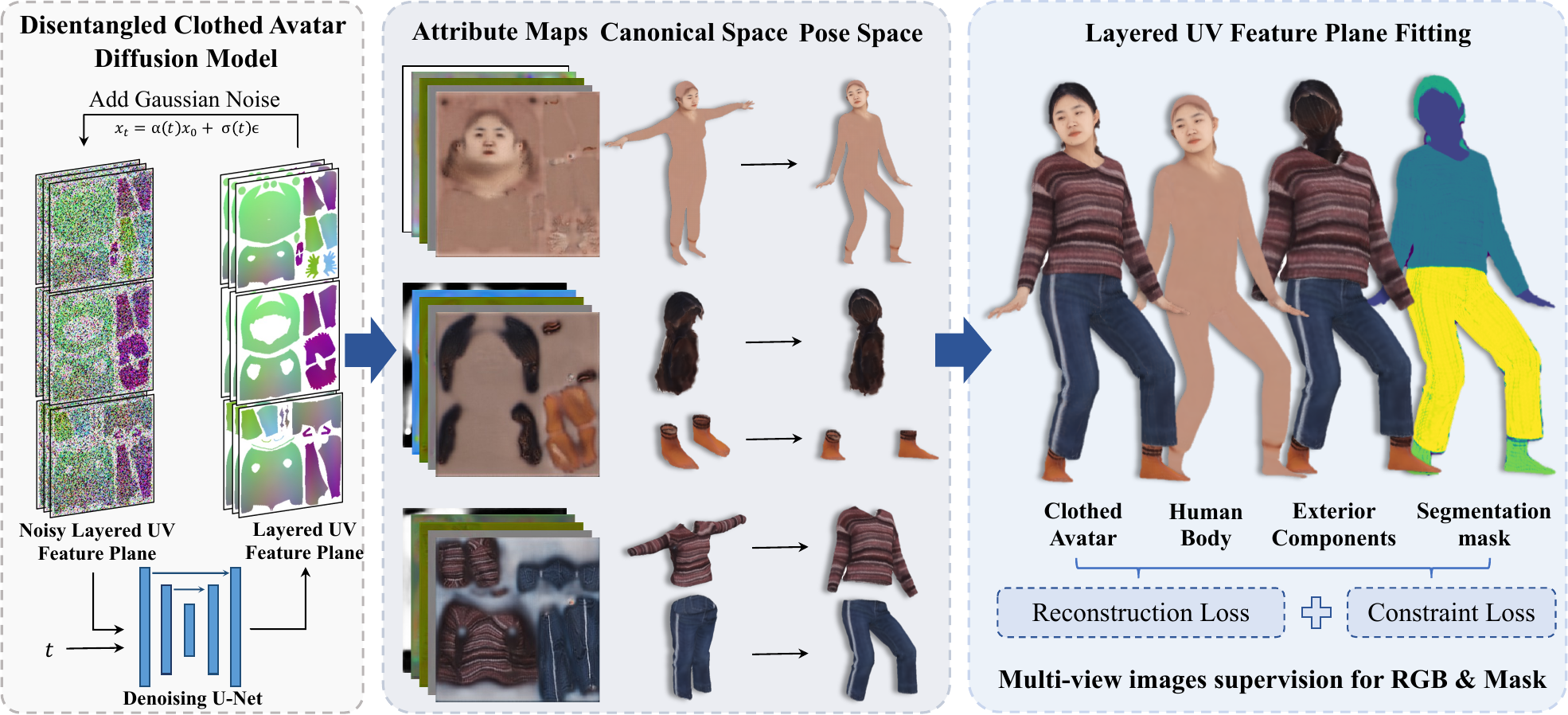}
   \caption{Method overview. LayerAvatar learns a feed-forward diffusion model to generate clothed avatars with each component disentangled. The clothed avatars are represented as layered UV feature plane where components are represented separately. After decoding the feature plane into attribute maps, we can extract 3D Gaussians from them through SMPL-X-based templates. Generated clothed avatars are then transformed into targeted pose space for further supervision. Reconstruction loss and constraint loss are both utilized to facilitate the disentanglement and handle the severe occlusion of human body layer.}
   \label{fig:method}
\end{figure*}
Inspired by general 3D object generation, many methods~\cite{bergman2022gnarf, chen2023primdiffusion, dong2023ag3d, zhang2023getavatar,noguchi2022unsupervised,hong2023eva3d} introduce 3D-aware GANs~\cite{Chan2022eg3d} and diffusion models~\cite{rombach2021sd} for clothed avatar generation. These methods first learn clothed avatars in canonical space and then animate them to posed space using a deformation module. 
Following EG3D~\cite{Chan2022eg3d}, some methods~\cite{zhang2022avatargen, zhang2023getavatar} apply triplane features to represent clothed avatars for higher quality and utilize inverse skinning for animation. AG3D~\cite{dong2023ag3d} introduces forward skinning technique~\cite{chen2021snarf,Chen2023fastsnarf} to achieve robust animation including loose clothing. On the other hand, Chupa~\cite{kim2023chupa} and AvatarPopup~\cite{kolotouros2024avatarpop} apply diffusion models to learn front and back view image pairs and then lift them to 3D space. 
Despite their impressive success, most of these methods represent clothed avatars as an entity and fail to disentangle the human body and clothes. Recently, some optimization-based methods~\cite{dong2024tela, gong2024laga} achieved success in generating cloth-disentangled avatars, however, the generating process takes hours to generate a single avatar.

\noindent \textbf{Compositional Avatar Representation.}
Instead of representing avatars~\cite{weng2022humannerf, yan2024dialoguenerf, moon2024expressive} as a single entity, some methods represent clothed avatars as a combination of multiple submodules. COAP~\cite{mihajlovic2022coap}, Spams~\cite{palafox2022spams} and DANBO~\cite{su2022danbo} consider human avatars as a composition of body parts, while EVA3D~\cite{hong2023eva3d} and ENARF-GAN~\cite{noguchi2022unsupervised} follow this trend and utilize multiple neural networks to represent different body parts of the digital avatar, achieving more efficient and detailed generation results. Several methods~\cite{abdal2023GSM, xu2023lsv, HumanLiff} represent clothed avatars as separate layers to enable the expressiveness of various topologies. However, these methods ignore the disentanglement of human body and clothes, which makes each submodule less physically meaningful. Recently, some works~\cite{Feng2022scarf, peng2024pica, lin2024layga, xiang2022dressing, zielonka25d3ga} disentangle the human body and clothes by representing each component separately. During the rendering process, these components are combined through various compositional rendering techniques. Most of these methods are designed to optimize a single digital avatar. In this paper, we propose a novel layered UV feature plane representation that is compatible with feed-forward generation framework.

\section{Method}
We propose LayerAvatar, a feed-forward generative method for disentangled clothed avatar generation. The overview of our method is illustrated in Fig.~\ref{fig:method}. 
We provide a brief introduction for prior knowledge in Sec.~\ref{subsec:pre}. 
To achieve disentangled clothed avatar generation, we propose a novel layered UV feature plane representation  (Sec.~\ref{subsec:rep}), which facilitates disentanglement and is compatible with current feed-forward generative pipelines. The clothed avatars are generated in canonical space and then deformed to targeted pose space via deformation module (Sec.~\ref{subsec:deform}). The training process is introduced in Sec.~\ref{subsec:train}. 

\subsection{Preliminary}
\label{subsec:pre}
\noindent \textbf{SMPL-X}~\cite{SMPL-X:2019} is an expressive parametric human model that can produce naked human meshes $ M(\mathbf{\beta}, \mathbf{\theta}, \mathbf{\psi})$ given shape parameter $\mathbf{\beta}$, pose parameter $\mathbf{\theta}$, and expression parameter $\mathbf{\psi}$. 
The producing process can be formulated as:
\begin{align}
\begin{split}
    T(\mathbf{\beta}, \mathbf{\theta}, \mathbf{\psi}) &= T_c+B_s(\mathbf{\beta};s)+B_e(\mathbf{\psi};e)+B_p(\mathbf{\theta};p), \\
    M(\mathbf{\beta}, \mathbf{\theta}, \mathbf{\psi}) &= LBS(T(\mathbf{\beta}, \mathbf{\theta}, \mathbf{\psi})), J(\mathbf{\beta}), \mathbf{\theta}, \mathcal{W}),
\end{split}
\end{align}
where a canonical human mesh $T$ is first calculated as a combination of the mean shape template $T_c$ and vertex displacements ($B_s(\mathbf{\beta};s)$, $B_e(\mathbf{\psi};e)$, $B_p(\mathbf{\theta};p)$) computed by the blend shapes $s$, $e$, $p$ and their corresponding pose, shape, and expression parameters. The body template $T$ is then deformed to the given pose by linear blend skinning(LBS) based on the skinning weights $\mathcal{W}$ and joint locations $J(\mathbf{\beta})$.

\noindent \textbf{3D Gaussians}~\cite{kerbl3Dgaussians} is a primitive-based explicit representation that combines the strengths of both previous explicit and implicit representations. It consists of a set of learnable 3D Gaussian primitive $\mathcal{G}_k$ where each contains five attributes: position $\boldsymbol{\mu}$, scaling matrix $\mathbf{S}$, rotation matrix $\mathbf{R}$, opacity $\alpha$, and color $\mathbf{c}$. In practice, we employ diagonal vector $\mathbf{s} \in \mathbb{R}^3$ and axis-angle $\mathbf{r} \in \mathbb{R}^3$ to represent $\mathbf{S}$ and $\mathbf{R}$ respectively. 3D Gaussians are represented as ellipses in 3D space defined by their position $\mu$ and covariance matrix $\mathbf{\Sigma}=\mathbf{R} \mathbf{S} \mathbf{S}^T \mathbf{R}^T$.
During the rendering process, these 3D Gaussians are projected to a 2D image plane where the pixel color $\mathbf{C}$ can be calculated as follows:
\begin{equation}
    \mathbf{C}=\sum_{i=1}^N \mathbf{c}_i \sigma_i \prod_{j=1}^{i-1}\left(1-\sigma_j\right),
\end{equation}
where $\mathbf{c}_i$ is the color of the $i$-th 3D Gaussian on the ray, and $\sigma_i$ is the blending weight calculated with the opacity $\alpha$.

\subsection{Layered UV Feature Plane Representation}
\label{subsec:rep}

Previous methods mainly represent avatars as a single entity or depend on optimization-based schemes. As a result, the generated results often have difficulty with editing or are slow to create, typically taking hours to generate a single subject. To address these limitations, we propose a layered UV feature plane representation that can separate components of clothed avatars and is compatible with fast, feed-forward generation pipelines. We employ 3D Gaussians as the base representation for efficient rendering along with easy animation and editing. 

To enable disentanglement, we consider clothed avatars as a composition of human body and exterior components. 
\begin{equation}
    \mathcal{G}_\text{avatar}=\{\mathcal{G}_\text{body}, \mathcal{G}_\text{top},\mathcal{G}_\text{bottom}, \mathcal{G}_\text{hair}, \mathcal{G}_\text{shoes}\}.
\end{equation}
Each component is represented as a set of Gaussian primitives $\mathcal{G}_i$ parameterized with five attributes: 3D position $\boldsymbol{\mu}_i \in \mathbb{R}^3$, opacity $\alpha_i \in \mathbb{R}$, rotation matrix $\mathbf{R}_i$ represented by axis angle $\mathbf{r}_i\in\mathbb{R}^3$, scale matrix $\mathbf{S}_i$ represented by diagonal vector $\mathbf{s}_i \in \mathbb{R}^3$ and rgb color $\mathbf{c}_i \in \mathbb{R}^3$. Inspired by existing works~\cite{hu2024gauhuman,zhang2024e3gen} that initialize 3D Gaussians by attaching them to SMPL parametric model for geometry and animation prior. We initialize the 3D Gaussians of each component by attaching them to self-designed templates based on SMPL-X. For each component, we design a template that maximizes coverage of the region where the component may exist. To enhance generation quality, all templates are subdivided to support densified Gaussian primitives. Then, we initialize the positions of 3D Gaussians as the center points of faces on the densified template mesh. And the initial rotations of 3D Gaussians are set as the tangent frame of the faces, which consists of the normal vector of that face, the direction vector of one edge and their cross product.

For compatibility with feed-forward generation pipelines, previous methods typically project 3D Gaussians into 2D space (\eg, multi-view image space~\cite{tang2024lgm, szymanowicz24splatter}, UV space~\cite{zhang2024e3gen}). Unlike these methods, which represent the entire subject as a single entity and then employ a post-processing step for disentanglement, we directly model components separately by mapping their templates into a three-layer UV space, combined with semantic labels. The first layer represents the innermost part of the human body, the second layer represents the hair and shoes, and the third layer represents top and bottom clothes. Following ~\cite{zhang2024e3gen}, all Gaussian attributes are stored as local UV features to enhance generation quality. The UV features plane is first split into two parts in a channel-wise manner and then decoded by two light-weight shared MLP decoders $\mathcal{D}_g$ and $\mathcal{D}_t$ separately. $\mathcal{D}_g$ decodes the geometry-related attributes: position offset $\Delta \boldsymbol\mu$ and opacity $\alpha$ of 3D Gaussians and $\mathcal{D}_t$ predicts texture-related attributes, color $\mathbf{c}$ and covariance-related rotation $\Delta \mathbf{r}$ and scale $\Delta \mathbf{s}$. Given the decoded attribute maps, we can extract the attributes for each 3D Gaussian $\mathcal{G}_i$ via bilinear interpolation. The opacity value $\alpha_i$ and color value $\mathbf{c}_i$ are obtained directly, and the other values are obtained via following formula based on their initial values:
\begin{equation}
    \boldsymbol{\mu}_i = \boldsymbol{\mu}^0_i + \Delta\boldsymbol{\mu}_i, \quad
    \mathbf{s}_i = \mathbf{s}^0_i \cdot \Delta\mathbf{s}_i, \quad
    \mathbf{r}_i = \mathbf{r}^0_i \cdot \Delta\mathbf{r}_i,
\end{equation}
where $\boldsymbol{\mu}^0_i$, $\mathbf{s}^0_i$, and $\mathbf{r}^0_i$ are initial position, scale, and rotation values for 3D Gaussians, respectively. $\Delta\boldsymbol{\mu}_i$, $\Delta\mathbf{s}_i$ and $\Delta\mathbf{r}_i$ are predicted residuals extracted from attribute maps. By collecting the attributes of 3D Gaussians with the same semantic label, we can obtain a canonical space representation of each component, and their composition forms the complete disentangled clothed digital avatar.

\subsection{Deformation}
\label{subsec:deform}
Benefiting from the SMPL-X-based templates, our method supports deformation in body shapes and novel poses including gestures and facial expressions. To support training with multiple subjects in various body shapes, we disentangle the body shape factor by defining all the templates in a canonical space with neutral body shapes. The neutral body shape avatar and its corresponding components can be transformed into targeted body shape space via the following warping process:
\begin{equation}
    \bar{\boldsymbol{\mu}} = \boldsymbol{\mu} + B_s(\mathbf{\beta}, s, \boldsymbol{\mu}),
\end{equation}
where $\bar{\boldsymbol{\mu}}$ represents the position of 3D Gaussians in the targeted $\mathbf{\beta}$ body shape space and $B_s(\mathbf{\beta}, s, \boldsymbol{\mu})$ are corresponding body shape related offsets extracted from the SMPL-X based templates via barycentric interpolation. We further add pose-dependent offsets $B_p(\mathbf{\theta}, p, \boldsymbol{\mu})$ and facial expression offsets $B_e(\mathbf{\psi}, e, \boldsymbol{\mu})$ in the same way to ensure accurate animation results.

The animation of a generated avatar from the canonical T-pose to an arbitrary target pose can be regarded as transforming the 3D Gaussian attributes. During the animation process, the opacity $\alpha$ and color $\mathbf{c}$ of 3D Gaussians remain unchanged. Therefore, we only discuss the transformation of position $\boldsymbol{\mu}$, rotation matrix $\mathbf{R}$ and scale matrix $\mathbf{S}$ in this section. Using the LBS function, we can transform the position $\bar{\boldsymbol{\mu}}$ of 3D Gaussians as:
\begin{equation}
    \boldsymbol{\mu}^{\prime}=\sum_{i=1}^{n_b} w_i \mathbf{B}_i \bar{\boldsymbol{\mu}},
\end{equation}
where $n_b$ represents the number of joints and $\mathbf{B}_i$ is the transformation matrix of the $i$-th joint. For 3D Gaussians on the innermost human body layer, the corresponding blend skinning weights $w$ are obtained directly from SMPL-X-based templates through barycentric interpolation, as these regions usually undergo minimal topology changes. For 3D Gaussians representing the exterior components, we follow ~\cite{dong2023ag3d} to extract the skinning weights from a pre-computed low-resolution volumetric field of fused skinning weights, which is more stable for points that deviate significantly from the original template. $\mathbf{T}=\sum_{i=1}^{n_b} w_i \mathbf{B}_i$ is the blended transformation matrix, and the rotation matrix $\mathbf{R}$ is updated via $\mathbf{R}^{\prime} = \mathbf{T}_{1: 3,1: 3} \mathbf{R}$, where $\mathbf{T}_{1: 3,1: 3}$ is the rotational part of $\mathbf{T}$. The scale matrix $\mathbf{S}$ is recalculated in the targeted pose space to fit deformed topology.

\subsection{Learning Disentangled Clothed Avatar}
\label{subsec:train}
To mitigate the impact of occlusion, we adopt a single-stage training scheme~\cite{ssdnerf}, which is more robust in occluded and sparse view situations. Specifically, the layered UV feature plane fitting and diffusion training process is conducted simultaneously, and the UV feature plane is jointly optimized by the fitting and diffusion loss. Similar to the SDS loss~\cite{poole2022dreamfusion}, the diffusion loss provides a diffusion prior for the UV feature plane, thereby facilitating the completion of unseen regions in the training images. 

\noindent\textbf{Layered UV Feature Plane Fitting.}
Given multi-view images, we optimize the layered UV feature plane and shared decoders to reconstruct avatars with disentangled components. The objective function can be divided into reconstruction and constraint part. The reconstruction loss $\mathcal{L}_{\text{recon}}$ can be formulated as follows:
\begin{equation}
    \mathcal{L}_{\text {recon}}=\lambda_{\text {color}} \cdot \mathcal{L}_{\text{color}}+\lambda_{\text {mask}}\cdot\mathcal{L}_{\text {mask}}+\lambda_{\text {per}} \cdot \mathcal{L}_{\text{per}}+\lambda_{\text {seg}} \cdot \mathcal{L}_{\text{seg}}.
\end{equation}

To achieve the disentanglement between exterior components and human body, we not only minimize the color loss $\mathcal{L}_{\text {color}}$ and mask loss $\mathcal{L}_{\text {mask}}$ on the overall rendering result, but also perform supervision on each component. Specifically, we first render the 3D Gaussians corresponding to each component separately to obtain multi-view images of each component. Then, inspired by Clothedreamer~\cite{liu2024clothedreamer}, instead of blending the rendering results of these components via estimated depth order to obtain the rendering results of clothed avatars, we directly render all the 3D Gaussians to alleviate artifacts caused by the blending process. The silhouette masks of each component and the clothed avatar are obtained similarly. The ground truth of silhouette masks is estimated based on the semantic segmentation results predicted by Sapiens~\cite{khirodkar2024sapiens}. We apply Huber loss~\cite{huber1992robust} for both $\mathcal{L}_{\text {color}}$ and $\mathcal{L}_{\text {mask}}$ following SCARF~\cite{Feng2022scarf}, due to its robustness to the estimated noisy segmentation results. To enhance the details of generated results, we also employ a perceptual loss $\mathcal{L}_\text{per}$~\cite{johnson2016perceptual} to minimize the difference between extracted features of rendered outputs and targeted images. 

Components in overlapping regions may learn inverted color or opacity values due to incorrect depth ordering. To address this, we render semantic segmentation maps of the clothed avatar by assigning the segmentation label of each Gaussian as its color. We then minimize the distance to the predicted semantic segmentation map using the Huber loss, $\mathcal{L}_\text{seg}$, to encourage accurate depth ordering.

Due to the severe occlusion of the inner human body layer, we apply constraints on the geometry and texture of the human body to obtain reasonable results. Since the human body is always within the exterior layer, we employ the following constraints:
\begin{equation}
    \mathcal{L}_{\text{maskin}}=\lambda_{\text {maskin}} \text{ReLU}\left(\mathcal{R}_m^b(\mathcal{G_\text{body}})-M_\text{fg}\right),
\end{equation}
where $\mathcal{R}_m^b(\mathcal{G_\text{body}})$ represents the rendered silhouette of the human body, and $M_\text{fg}$ is the estimated foreground silhouette mask. When rendering $\mathcal{R}_m^b(\mathcal{G_\text{body}})$, we detach the opacity values and set them to 1, only optimizing offset and covariance-related attributes, preventing the model from minimizing the loss via decreasing opacity on the boundary. Utilizing the prior that the occluded skin color should be similar to the color of hands, we introduce the following texture constraint:
\begin{equation}
    \mathcal{L}_{\text {skin }}=\lambda_{\text {skin}} \left(M_\text{oc} \odot\left(\mathcal{R}_m^b(\mathcal{G}_\text{body})-\mathbf{C}_{\text{skin}}\right)\right),
\end{equation}
where $\mathbf{C}_{\text{skin}}$ is the average color of pixels in the hands region and $M_\text{oc}$ is the mask of the occluded region. Other regularization terms are as follows:
\begin{equation}
    \mathcal{L}_\text{reg} = \lambda_\text{offset}\mathcal{L}_\text{offset} + \lambda_\text{smooth}\mathcal{L}_\text{smooth}.
\end{equation}
$\mathcal{L}_\text{offset}=\|\Delta_{\mu}\|^2$ constrains the offset from being extremely large. $\mathcal{L}_\text{smooth}$ is the total variational (TV) loss, which is used to minimize the average $L_2$ distance between neighboring pixels on attribute maps. This regularization term encourages smooth transitions between the neighboring attributes (\eg offsets, rotation, and opacity), promoting the generation of reasonable texture and geometry surface.

\noindent\textbf{Disentangled Clothed Avatar Diffusion Model.}
To generate disentangled clothed avatars, we train a diffusion model that maps Gaussian noise to the layered UV latent space. Since diffusion models generally perform better on inputs with low channel dimensions~\cite{rombach2022high}, we concatenate our layered UV feature plane across widths instead of stacking them across channels. During training, we first obtain destructed UV feature plane $\boldsymbol{x}_t$ by adding Gaussian noise $\boldsymbol{\epsilon} \sim \mathcal{N}(0, \mathbf{I})$ to the layered UV feature plane $\boldsymbol{x}_0$ according to following noise schedule:
\begin{equation}
    \boldsymbol{x}_t:=\alpha(t) \boldsymbol{x}_0+\sigma(t) \boldsymbol{\epsilon},
\end{equation}
where $\alpha(t)$ and $\sigma(t)$ are predefined functions that control the intensity of added noise, and $t$ is the time step in the range of [0, 1]. To stabilize and accelerate training, we use v-prediction proposed in ~\cite{salimans2022progressive} to train the denoising UNet. The objective function for the diffusion model is:
\begin{equation}
            \mathcal{L}_{\text{diff}} =\underset{t,\boldsymbol{x}_0, \boldsymbol{\epsilon}}{\mathbb{E}}\left[\frac{1}{2} w(t)\left\|\ \hat{\boldsymbol{x}}_0-\boldsymbol{x}_0\right\|^2\right],
\end{equation}
where $w(t) =\left(\alpha(t) / \sigma(t)\right)^{2 \omega}$, the $\omega$ is set to 0.5 following ~\cite{ssdnerf}. $\mathcal{L}_{\text{diff}}$ is used to update not only the parameters of denoising UNet, but also the UV feature plane. It promotes the UV feature plane to adapt to the learned latent space, thereby providing priors for occluded regions.
\section{Experiments}
\begin{table*}
\centering
  \subfloat[
    Holistic Generation Quality. 
\label{tab:full_generation_quality}
]{
\begin{minipage}{0.21\linewidth}{\begin{center}
\tablestyle{4pt}{1.05}
\begin{tabular}{lc}
\toprule
Methods & FID$\downarrow$ \\
\midrule
EVA3D~\cite{hong2023eva3d} & 124.54$\dag$ \\
StructLDM~\cite{hu2024structldm} & 25.22$\dag$ \\
E3Gen~\cite{zhang2024e3gen} & 15.78$\star$ \\
\textbf{Ours} & \textbf{12.50} \\
\bottomrule
\end{tabular}
\end{center}}\end{minipage}
}
\hspace{1em}
\subfloat[
    Layer-wise Generation Quality. 
\label{tab:layer_generation_quality}
]{
\begin{minipage}{0.29\linewidth}{\begin{center}
\tablestyle{4pt}{1.05}
\begin{tabular}{lcccc}
\toprule
Methods & FID$\downarrow$ & L-PSNR$\uparrow$\\
\midrule
EVA3D~\cite{hong2023eva3d} & 61.58* & $<$20* \\
Rodin~\cite{wang2023rodin} & 56.57* & 18.12* \\
HumanLiff~\cite{HumanLiff} & 54.39* & 28.57* \\
\textbf{Ours} & \textbf{17.37} & \textbf{$>$40} \\
\bottomrule
\end{tabular}
\end{center}}\end{minipage}
}
\hspace{1em}
\subfloat[
    Component Quality and Disentanglement Evaluation. 
\label{tab:compo_generation_quality}
]{
\centering
\begin{minipage}{0.37\linewidth}{\begin{center}
\tablestyle{4pt}{1.05}
\begin{tabular}{lccc}
\toprule
Methods & Disentanglement$\uparrow$ & Quality$\uparrow$ & Time$\downarrow$ \\
\midrule
SO-SMPL~\cite{wang2023disentangled} & 13.02 & 3.57/8.57 & 5h \\
LAGA~\cite{gong2024laga} & 9.77 & 5.00/1.43 & 1.5h \\
TELA~\cite{dong2024tela} & 19.07 & 14.29/8.57 & 6h\\
\textbf{Ours} & 58.14 & 77.14/81.43 & 2s\\
\bottomrule
\end{tabular}
\end{center}}\end{minipage}
}
\caption{Quantitative comparison. Our method outperforms other methods in both holistic and layer-wise generation quality, as well as disentanglement capability. Due to the minor difference between the two masked layers, the L-PSNR value of our method is so large that we use $>$ 40 to express it. For component comparison, we show the quality preference in \textit{component quality preference/overall quality preference} format. $*$, $\dag$ and $\star$ denote results adopted from HumanLiff~\cite{HumanLiff}, StructLDM~\cite{hu2024structldm} and E3Gen~\cite{zhang2024e3gen} respectively.}
\label{tab:compare}
\vspace{-1mm}
\end{table*}
\begin{figure*}
  \centering
   \includegraphics[width=\linewidth]{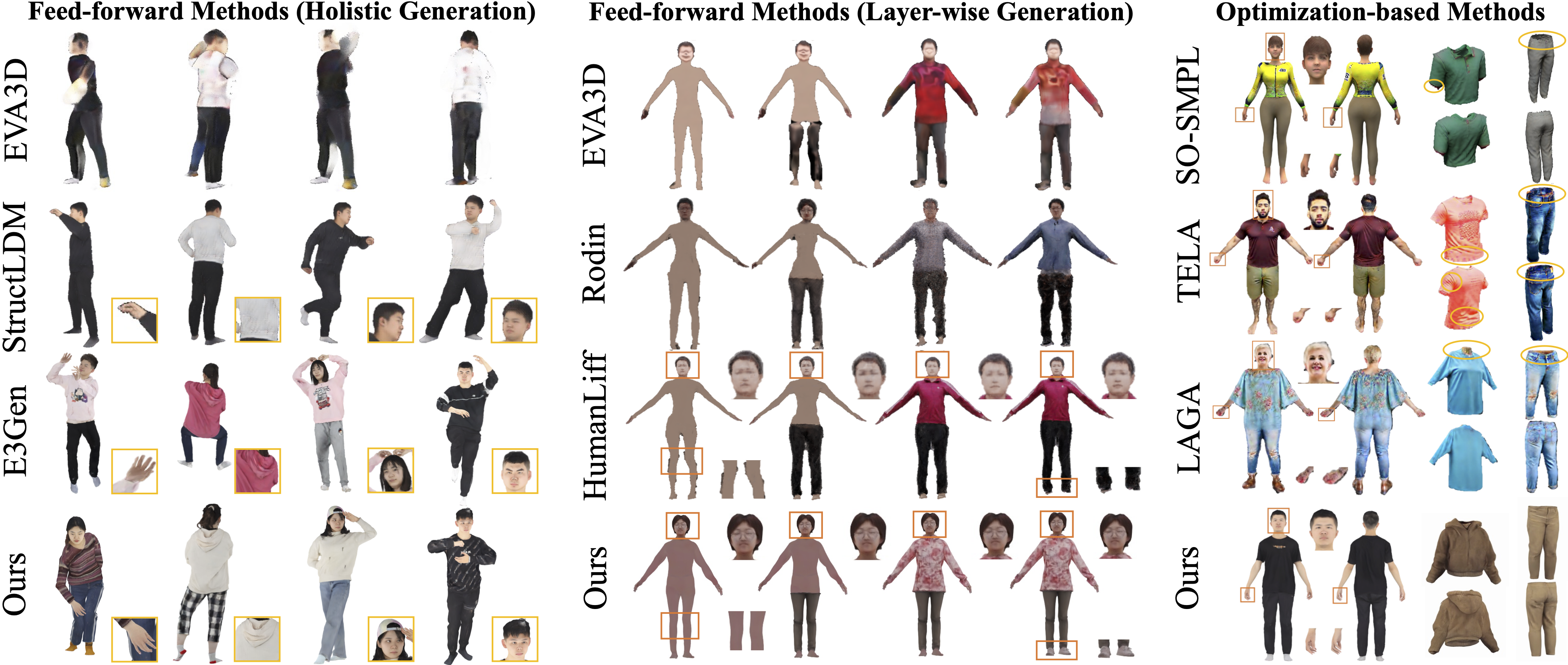}
   \caption{Qualitative comparison. The left part demonstrates our generation results on THuman2.0 dataset~\cite{tao2021function4d}. The middle part illustrates layer-wise generated results on Tightcap dataset~\cite{chen2021tightcap}. Our method can generate more reasonable human bodies under severe occluded scenarios. The right part exhibits our component generation results compared with optimization-based methods.}
   \label{fig:compare}
   \vspace{-3mm}
\end{figure*}
\noindent\textbf{Baselines.} To evaluate holistic generation quality, we compare our method with the state-of-the-art methods of animatable avatar generation (EVA3D~\cite{hong2023eva3d}, StructLDM~\cite{hu2024structldm}, and E3Gen~\cite{zhang2024e3gen}) on THuman2.0~\cite{tao2021function4d} dataset. We then evaluate the decomposition capability of our method by comparing the layer-wise generation results with the disentanglement-related method  HumanLiff~\cite{HumanLiff} on Tightcap~\cite{chen2021tightcap} dataset. And we further evaluate the component generation quality against optimization-based methods: LAGA~\cite{gong2024laga}, SO-SMPL~\cite{wang2023disentangled}, and TELA~\cite{dong2024tela}. Our method is trained on a composite dataset including CustomHuman~\cite{ho2023custom}, THuman2.0~\cite{tao2021function4d}, and THuman2.1~\cite{tao2021function4d}.

\noindent\textbf{Metrics.}
For holistic generation quality evaluation, we utilize FID~\cite{heusel2017fid} following previous works~\cite{Chan2022eg3d,hu2024structldm,zhang2024e3gen}. For layer-wise generation results, we adopt FID for overall generation quality evaluation and L-PSNR~\cite{HumanLiff} for disentanglement capability evaluation, which calculates the PSNR between two layers with the adding component masked. Additionally, we conduct a user study with 20 participants to compare the generation quality and disentanglement quality of our methods with optimization-based methods.

\noindent\textbf{Dataset.}
For THuman2.0~\cite{tao2021function4d}, we sample 500 scans and render each from 54 camera views to obtain multi-view images. Then, we employ Sapiens~\cite{khirodkar2024sapiens} to estimate segmentation masks for these images, from which per-component silhouette masks can be extracted. This approach enables our method to learn directly from multi-view images without requiring separate meshes of each component, thereby simplifying the data collection process.
Tightcap~\cite{chen2021tightcap} contains 3D scans with separate meshes for cloth and shoe, which facilitates direct rendering of multi-view images and silhouette masks for each component. For a fair comparison, we use the preprocessed version provided by HumanLiff~\cite{HumanLiff}, which selects 107 samples from Tightcap. Each sample is rendered from 158 camera views, providing images and silhouette masks for both the entire avatar and each component.
We also construct a composite dataset consisting of 1954 selected scans from THuman2.0~\cite{tao2021function4d}, THuman2.1~\cite{tao2021function4d}, and CustomHuman~\cite{ho2023custom} datasets. Each scan is processed in the same way as the THuman2.0 data. Additionally, all SMPL-X parametric models are standardized to neutral gender with refinement in body shape parameters to ensure the body layer fits underneath the cloth surface.

\subsection{Evaluation of Generation Quality and Disentanglement Capability}
\noindent\textbf{Disentanglement and Animation Capacities.}
The disentanglement and animation results are shown in Fig.~\ref{fig:teaser}. Our method successfully generates clothed avatars with full disentanglement of components such as hair, shoes, and clothes from the human body. Moreover, our method reconstructs the inner body layer with reasonable geometry and texture under severe occlusion. Benefiting from the SMPL-X-based templates within our layered Gaussian-based UV feature plane representation, each component can be easily deformed into novel poses. 

\noindent\textbf{Comparisons.}
For holistic generation quality, we compare our method against representative feed-forward diffusion and 3D-aware GAN pipelines. The quantitative comparison results are shown in Tab.~\ref{tab:full_generation_quality}. Our method outperforms all baselines on THuman2.0~\cite{tao2021function4d} dataset, achieving the best FID score. The visual comparisons shown in Fig.~\ref{fig:compare} further enhance the superiority of our method. EVA3D~\cite{hong2023eva3d} struggles with generating digital avatars with plausible texture and geometry. Compared to StructLDM~\cite{hu2024structldm} and E3Gen~\cite{zhang2024e3gen}, our method exhibits finer details, such as distinctive fingers, realistic cloth wrinkles, and intricate faces. 

To evaluate disentanglement capability, we compare layer-wise generation results and disentangled avatar generation results of our method with state-of-the-art methods. For layer-wise generation, our method surpasses other methods in both FID and L-PSNR, as shown in Tab.~\ref{tab:layer_generation_quality}, which indicates that our method achieves higher holistic generation results and can generate disentangled avatars without facing the identity shifting problem. The elimination of identity shifting demonstrates that our method achieves full disentanglement of components, since entangled components will result in shifting problems. As shown in Fig.~\ref{fig:compare}, Rodin~\cite{wang2023rodin} struggles to maintain identity consistency during the generation process. Although HumanLiff~\cite{HumanLiff} achieves correct layer-wise generation, it produces less smooth body geometry and exhibits subtle identity shifting, shown by the enlarged face region. Our method can generate each layer without shifting problems, demonstrating more complete disentanglement.

For disentangled avatar generation, we compare our method with state-of-the-art optimization-based methods. As shown in Tab.~\ref{tab:compo_generation_quality}, our method generates clothed avatars in significantly less time and achieves a higher user preference. Qualitative comparisons are illustrated in Fig.~\ref{fig:compare}. SO-SMPL~\cite{wang2023disentangled} generates unrealistic, cartoonish colors and exhibits unnatural sawtooth on the border of generated components (highlighted by orange circles). LAGA~\cite{gong2024laga} and TELA~\cite{dong2024tela} struggle with blurry fingers, oversaturated colors, and incomplete disentanglement from the body layer. In contrast, our method generates fully disentangled avatars with distinctive fingers and realistic color.

\subsection{Ablation Study}
\noindent\textbf{Layered \vs Single-Layer Representation.}
\begin{table}
  \centering
  \setlength{\tabcolsep}{7mm}{
  \begin{tabular}{lcc}
    \toprule
    Methods & FID$\downarrow$ & KID $\downarrow$ \\
    \midrule
    Two-stage & 19.06  & 15.40 \\
    Pipeline (SLMO) & 30.83 & 30.18 \\
    Pipeline (SLMN) & 28.95 & 29.21 \\
    \textbf{Full pipeline} & \textbf{12.50} & \textbf{9.39} \\
    \bottomrule
  \end{tabular}}
  \caption{Ablation study on THuman2.0 Dataset. The full pipeline outperforms baselines on both FID and KID by a large margin.}
  \vspace{-5mm}
  \label{tab:ablation}
\end{table}
\begin{figure}
  \centering
   \includegraphics[width=\linewidth]{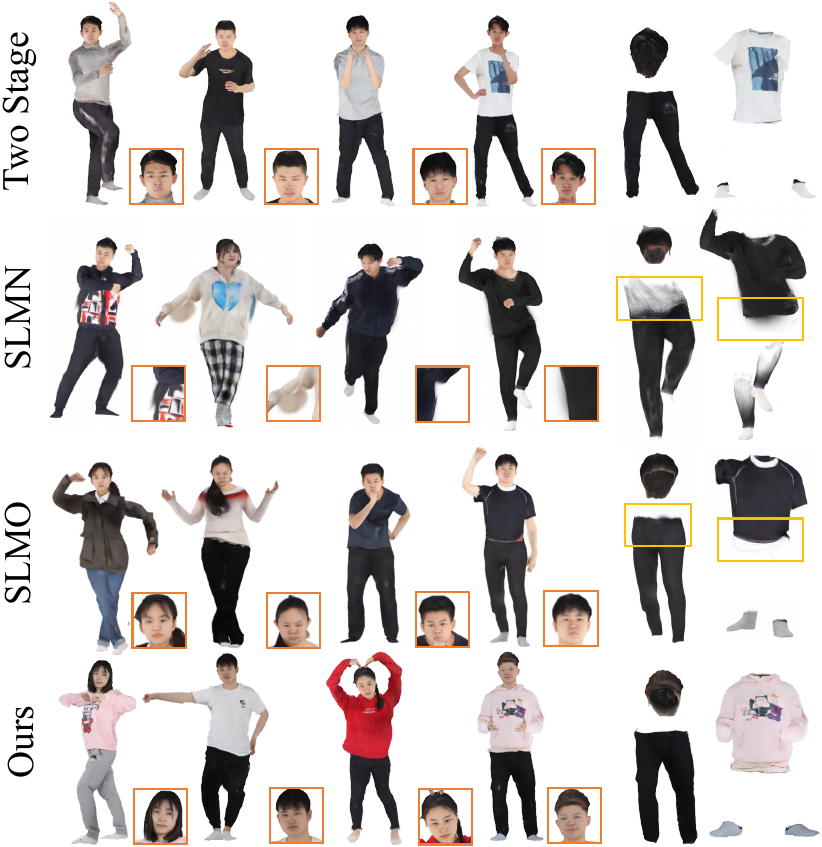}
   \caption{Ablation study on THuman2.0 Dataset. Comparing randomly generated avatars and decomposition results, our method generates avatars with higher quality and better disentanglement.}
   \label{fig:abl}
    \vspace{-4mm}
\end{figure}

\begin{figure}
    \centering
    \includegraphics[width=\linewidth]{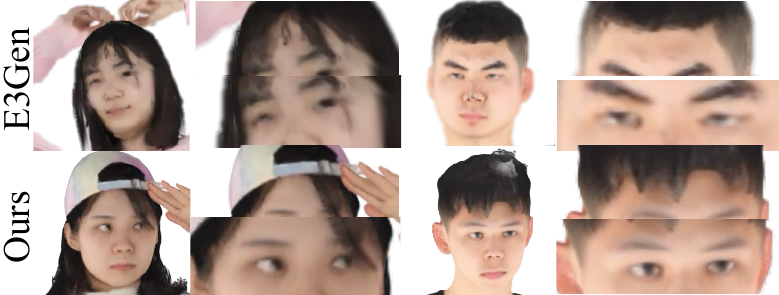}
    \caption{Zoom in comparison between E3Gen and LayerAvatar. Compared to single-layer representation, layered representation exhibits detailed faces and clear boundaries between components.}
    \label{fig:abl_com}
\end{figure}

To demonstrate the effectiveness of our layered UV feature plane, we design two baselines that use single-layer UV feature plane to generate disentangled clothed avatars. One baseline representation is SLMO (single-layer-multi-output), which utilizes a single pair of geometry and texture decoders ($\mathcal{D}_g$ and $\mathcal{D}_t$) to predict attributes for all components. Another one is SLMN (single-layer multi-network), which employs different decoders to predict the attributes of Gaussian primitives for each component. Since they only contain a single-layer UV feature plane, attributes of different components will share one feature if they are initialized in the same location. As shown in Fig.~\ref{fig:abl}, Fig.\ref{fig:abl_com} and Tab.~\ref{tab:ablation}, single-layer UV representations generate clothed avatars with lower quality both quantitatively and qualitatively. This is because our layered UV feature plane representation provides each component with an optimized UV distribution and independent feature space. Without the layered UV plane, which provides distinctive features for components in overlapping and neighboring regions, single-layer representations struggle to disentangle components, leading to blurred boundaries between components.

\noindent\textbf{Single Stage \vs Two Stage.}
We also compare the single-stage training scheme with the commonly used two-stage training scheme, shown in Tab.~\ref{tab:ablation} and Fig.~\ref{fig:abl}. The single-stage training can generate avatars with finer details, such as intricate faces.

\subsection{Applications}
\noindent\textbf{Component Transfer.}
\begin{figure}[t]
  \centering
   \includegraphics[width=\linewidth]{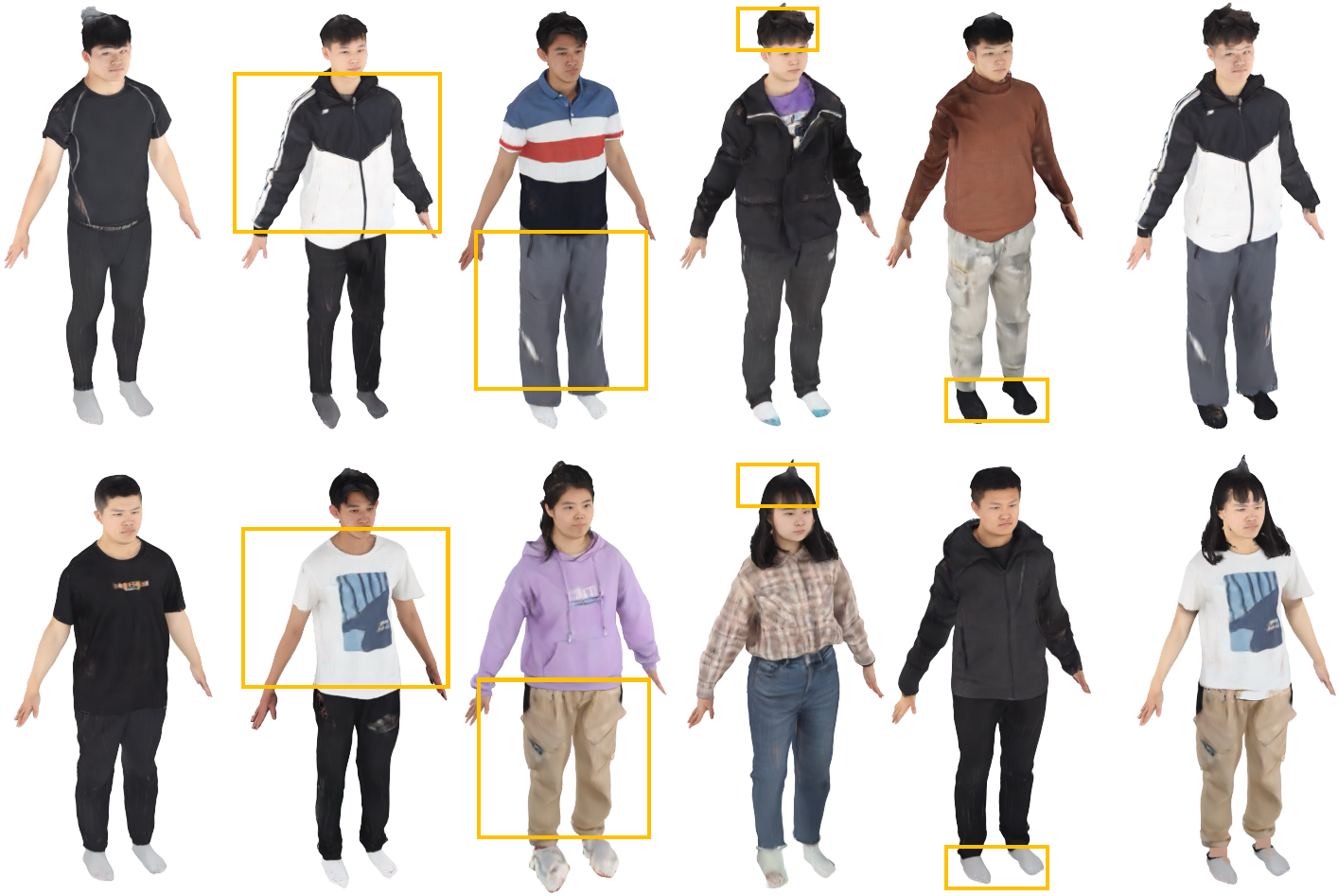}
   \caption{Component transfer. Given the generated avatars in the first column, we can transfer the upper-clothes, pants, hair, and shoes of the avatars in the second to fifth column to them. The results are shown in the rightmost column.}
   \label{fig:transfer}
    \vspace{-2mm}
\end{figure}
We further explore applications such as component transfer. Thanks to the component disentanglement and shared structure provided by our method, we can directly transfer clothes and other components between generated samples. The component transfer results are shown in Fig.~\ref{fig:transfer}. Our method can accurately transfer components across various body shapes while maintaining high-quality details of the transferred items.
\section{Conclusion}
In this paper, we propose LayerAvatar, a novel feed-forward diffusion-based method for generating component-disentangled clothed avatars. Our method proposes a layered UV feature plane representation, which organizes 3D Gaussians into different layers, each corresponding to specific components of clothed avatars (e.g., body, hair, clothing). Leveraging this representation, we train a single-stage diffusion model to generate each feature plane, enabling the generation of fully disentangled clothed avatars. To ensure complete component disentanglement, we incorporate constraint terms into the model. Extensive experiments validate the superiority of LayerAvatar in generating fully disentangled clothed avatars and its effectiveness in component transfer tasks. Additional limitations and implementation details are discussed in the supplementary material.

\PAR{Acknowledgments.}
This work was supported in part by NSFC (62201342), and Shanghai Municipal Science and Technology Major Project (2021SHZDZX0102).

{
    \small
    \bibliographystyle{ieeenat_fullname}
    \bibliography{main}
}

\clearpage
\setcounter{page}{1}
\maketitlesupplementary
\renewcommand{\thefigure}{\Alph{figure}}
\renewcommand{\thetable}{\Alph{table}}

This is the supplementary material for \textit{Disentangled Clothed Avatar Generation via Layered Representation}. A video (Sec~\ref{sec:video}) is included summarizing our method and exhibiting more visualization results. We introduce the implementation details of our method in Sec~\ref{sec:imp}. 
Additional experimental results and limitation discussions are provided in Sec~\ref{sec:limit}. 

\section{Supplementary Video}
\label{sec:video}
We provide a supplementary video for quick understanding of our method. The video includes: 
\begin{itemize} 
    \item A brief introduction of our method; 
    \item Results of unconditional generation and decomposition; 
    \item Results of novel pose animation; 
    \item Results of component transfer.
\end{itemize}

\section{Implementation Details}
\label{sec:imp}
\subsection{Network Architecture}
\begin{figure}
    \centering
    \includegraphics[width=\linewidth]{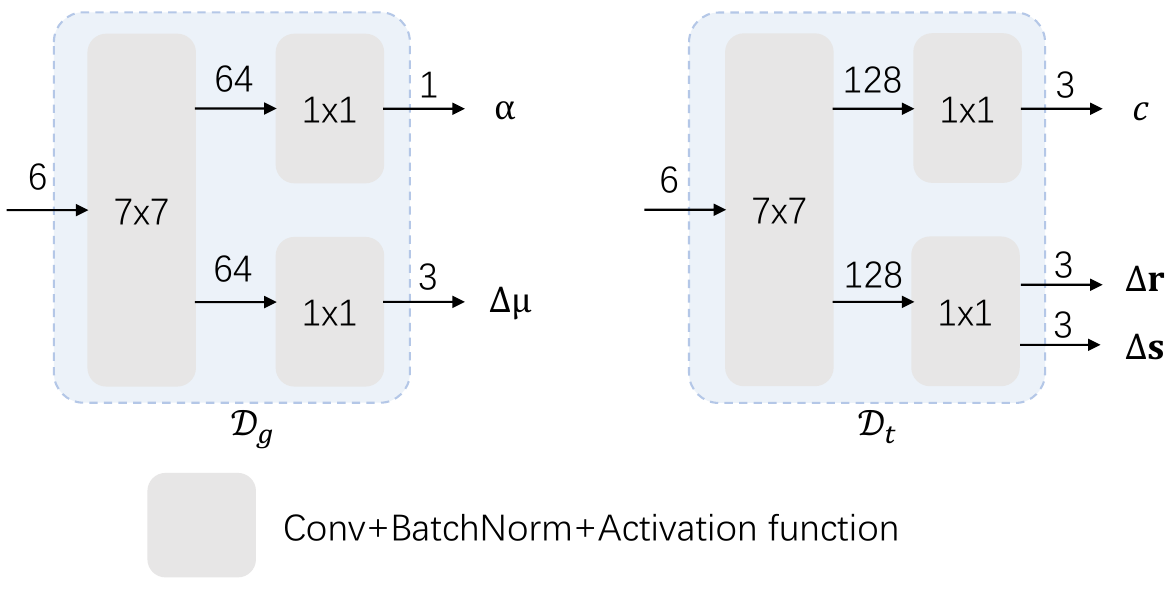}
    \caption{Network Architecture of Shared Decoder. The number above the arrow represents the input and output channels of each block. Each block consists of one 2D convolutional layer, one batchnorm layer, and one activation function. The number on the block is the kernel size of the convolutional layer. We utilize SiLU as the activation function and for the last layer, sigmoid is used as the activation function except for the $\Delta\mu$ output.}
    \label{fig:netarch}
\end{figure}

\noindent\textbf{Layered UV Feature Plane and Shared Decoders.}
The size of layered UV feature plane is $12\times128\times384$, where we concatenate the three-layer Gaussian-based UV feature plane width-wise instead of channel-wise following ~\cite{wang2023rodin}. Two shared decoders $\mathcal{D}_g$ and $\mathcal{D}_t$ are utilized to decode the UV feature plane to attribute maps, where attribute (position $\mu$, opacity $\alpha$, rotation $\mathbf{r}$, scale $\mathbf{s}$, color $\mathbf{c}$) of 3D Gaussians~\cite{kerbl3Dgaussians} can be extracted via bilinear interpolation. $\mathcal{D}_g$ predict geometry-related attributes (position and opacity) while $\mathcal{D}_t$ outputs texture-related attributes (color, rotation and scale). The architecture of these two shared decoders is shown in Fig.~\ref{fig:netarch}.
$\mathcal{D}_g$ and $\mathcal{D}_t$ both are shallow decoder with two layers. For the first layer, we apply SiLU~\cite{elfwing2018silu} as the activation function, while for the last layer, Sigmoid is utilized except for the offset $\Delta \mu$ prediction layer. No activation function is utilized for the offset prediction layer. Both the offset prediction layer and covariance prediction layer (predict $\Delta \mathbf{r}$ and $\Delta \mathbf{s}$) are initialized with weight conform to $\mathcal{U}(-1\times 10^{-5}, +1\times 10^{-1})$. Biases are initialized to be 0.

\noindent\textbf{Denoising UNet.}
Following~\cite{rombach2021sd}, the denoising UNet~\cite{ronnebergerunet} has UNet architecture with attention modules. The network architecture is shown in Tab.~\ref{tab:unet}. 
\begin{figure}
    \centering
    \includegraphics[width=\linewidth]{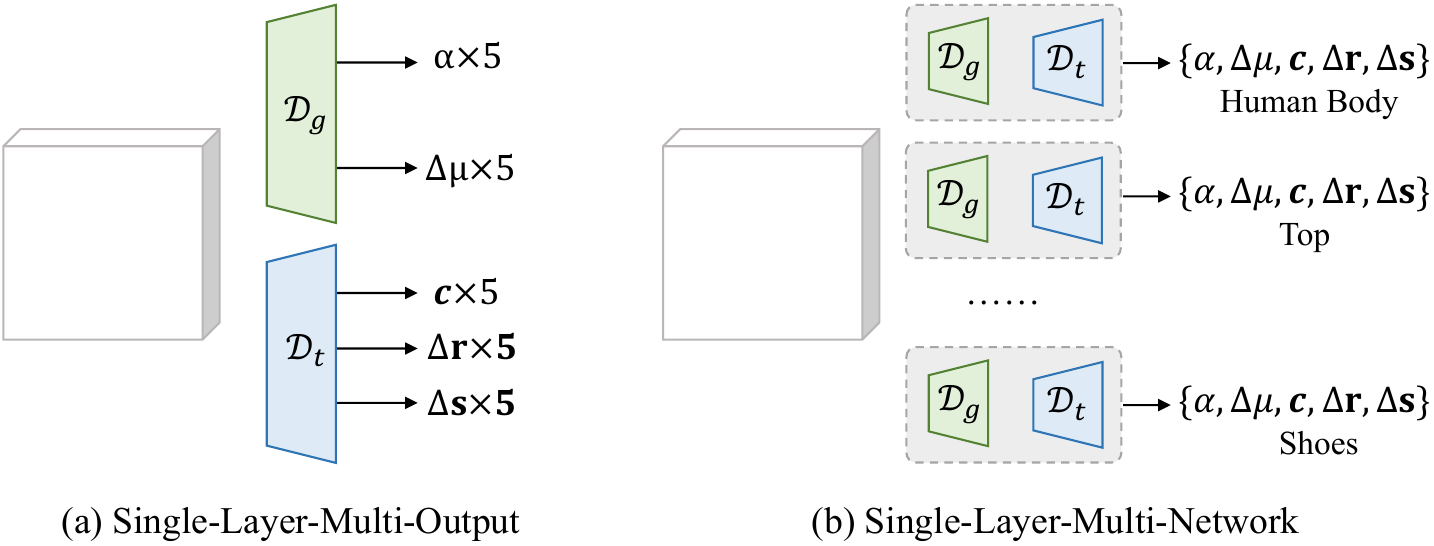}
    \caption{Structure of Single Layer representations. We design two kinds of single-layer representations for ablation study.}
    \label{fig:single_layer}
\end{figure}
\begin{table}
    \centering
    \caption{Model Configuration of denoising UNet.}
     \setlength{\tabcolsep}{4.0mm}
    {\begin{tabular}{lc}
    \toprule
    Key & Value \\
    \midrule
       Number of timesteps  &  1000\\
       Noise configuration  & Linear\\
       Input size & $128\times384\times12$\\
       Base channel & 128\\
       Channels configuration & [0.5, 1, 2, 2, 4, 4] \\
       Resblocks per downsample & 2 \\
       dropout & 0 \\
       Number of attention heads & 4\\
       Attention resolution & [16, 8, 4] \\
     \bottomrule
    \end{tabular}}
    \label{tab:unet}
    \vspace{-3mm}
\end{table}

\subsection{Training Details}
The whole training framework is built upon Pytorch. We utilize Adam~\cite{kingma2014adam} as the optimizer. For shared decoder and denoising UNet, the learning rate is $1\times10^{-4}$ and $1\times10^{-4}$ separately. The learning rate for UV feature plane is $0.04$. All the UV feature planes are first normalized through a Tanh function and then sent into the shared decoder. The total batch size is 4 scenes per GPU. During one iteration, we select 2 views of each scene for supervision. The whole training process takes 6 days on two RTX 3090 GPUs. The loss weight for each fitting loss is as follows:
$\lambda_\text{color}=18$, $\lambda_\text{mask}=9$, $\lambda_\text{per}=0.05$, $\lambda_\text{seg}=9$, $\lambda_\text{maskin}=5$, $\lambda_\text{skin}=0.5$,
$\lambda_\text{offset}=5$, and $\lambda_\text{smooth}=0.5$.
For ablation study, we propose single-layer representations with a size of $128\times128\times12$, the whole architecture is shown in Fig.~\ref{fig:single_layer}. Other parameters are the same as layered UV feature plane.

\subsection{Data Preprocessing}
For Tightcap~\cite{chen2021tightcap}, we convert the estimated SMPL~\cite{SMPL:2015} parameters to SMPLX~\cite{SMPL-X:2019} to fit our template. Masks for each component (top, bottom, shoes) are provided by the dataset. Since the dataset does not separate hair from body, we combine hair and body components during training. For THuman2.0, THuman2.1~\cite{tao2021function4d} and CustomHuman~\cite{ho2023custom}, we first optimize the SMPL-X parameters to make them underneath the clothing layer. We then obtain the segmentation maps for each view through Sapiens~\cite{khirodkar2024sapiens} estimation and merge the original 20 segmentation labels into 5 (hair, body, top, bottom, and shoes).

\section{Limitations and Discussions}
\label{sec:limit}
\subsection{Limitations}
(1) Due to the segmentation-map-based supervision and SMPLX-based templates, the performance of our method is affected by the accuracy of the estimated segmentation map and SMPLX parameters. Eliminating inaccurate segmentation results as in ~\cite{jiang2024multiply} and optimizing SMPLX parameters during training can help mitigate the impact.

\noindent(2) The collision between body and clothing is a long-existing problem when representing the human body and clothing separately. Introducing post-processing to optimize the position of 3G Gaussians on clothing and body via collision-avoiding loss might eliminate this problem. 

\noindent(3) Animation of loose clothing is prone to artifacts. Using video data instead of multi-view images or introducing pre-trained networks for cloth deformation might help.

\noindent(4) While our method effectively disentangles core components, a promising direction for future research is to extend it to handle general accessories (e.g., glasses, bags) by incorporating additional template layers or adopting other representations.

\noindent(5) 
Introducing physical constraints by converting current representation to meshes~\cite{korosteleva2022neuraltailor} or combining UV feature plane with sewing pattern similar to GarmageNet~\cite{li2025garmagenet} would be interesting future work.

\subsection{Additional Experiments}
\begin{figure}[h]
    \centering
    \includegraphics[width=\linewidth]{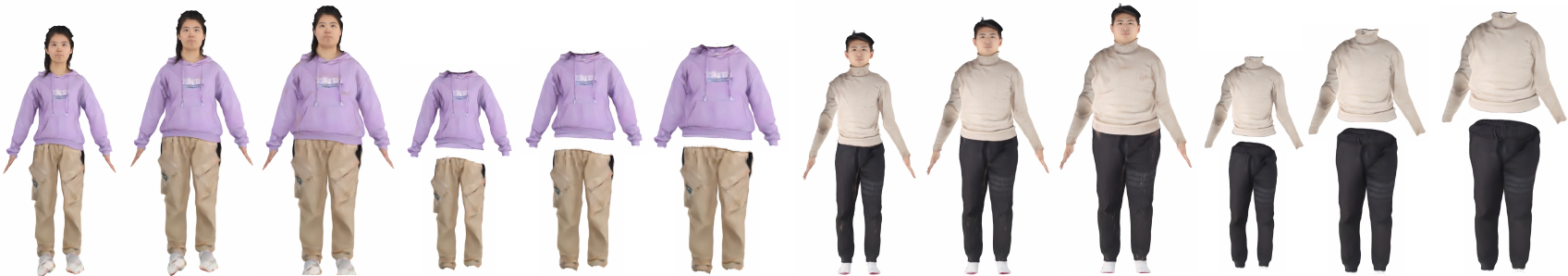}
    \caption{Clothing Adaptation to Body Shape}
    \label{fig:bodyshape}
\end{figure}

\noindent\textbf{Shape Adaptation.} Benefiting from the SMPL-X-based template of our representation, we can adapt clothing to various body shapes as shown in Fig.~\ref{fig:bodyshape} which facilitates seamless transfer of components among subjects.

\begin{figure}[h]
    \centering
    \includegraphics[width=\linewidth]{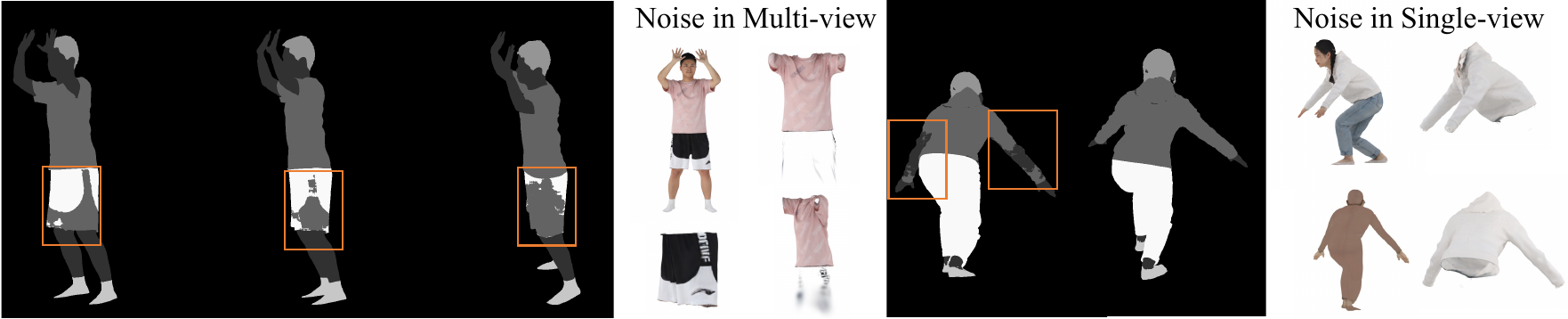}
    \caption{Robustness to noisy semantic masks.}
    \label{fig:noise}
\end{figure}

\noindent\textbf{Sensitivity on noisy semantic masks.}
Our method is robust to noisy masks thanks to multi-view supervision, which can correct errors present in individual views. During training, we filter out samples with obvious noise in over three consistent views. As shown in Fig.~\ref{fig:noise}, our method can handle noise that does not persist across multiple views.

\begin{figure}[h]
    \centering
    \includegraphics[width=\linewidth]{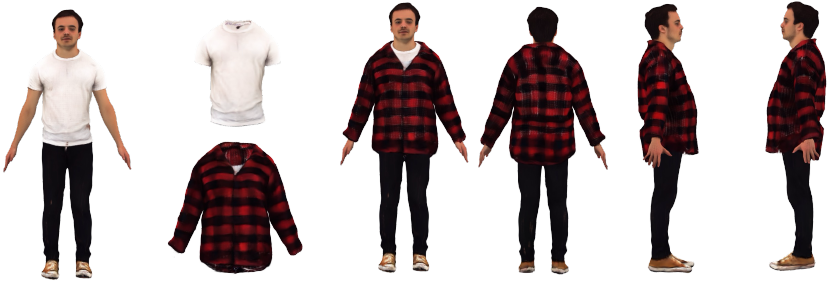}
    \caption{Multi-layer garment composition.}
    \label{fig:multigarment}
\end{figure}

\noindent\textbf{Extension to multi-layer outfits.} As shown in Fig.~\ref{fig:multigarment}, our method can composite multi-layer clothes to enable more composition.

\begin{figure}[h]
    \centering
    \includegraphics[width=\linewidth]{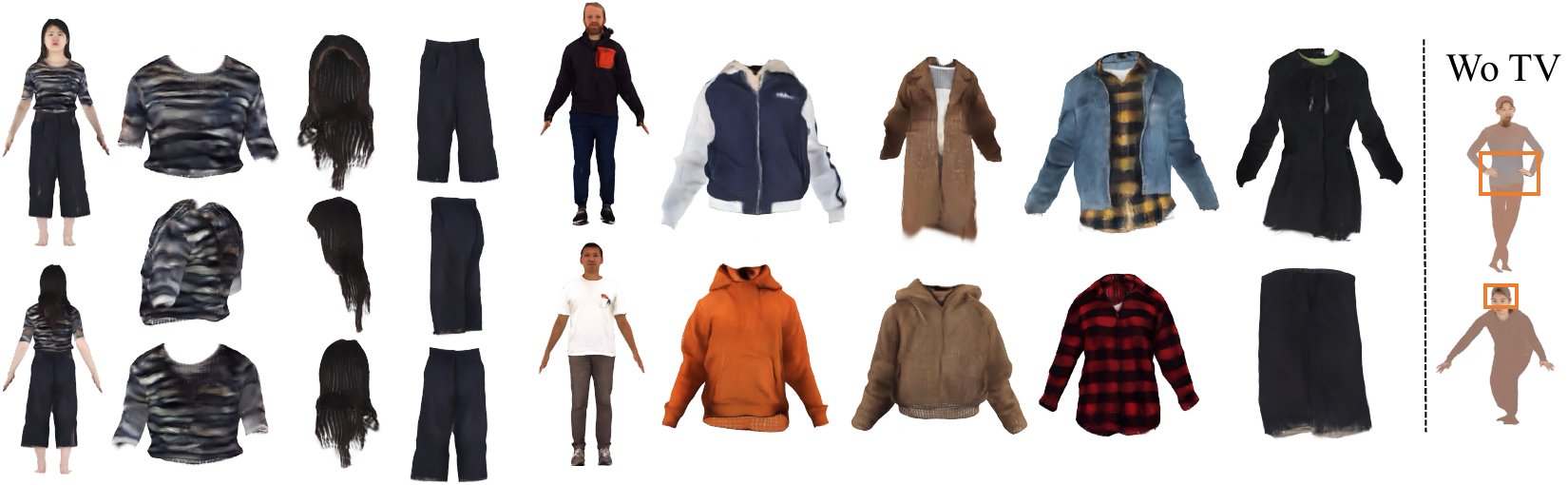}
    \caption{More diverse generation results.}
    \label{fig:extend}
\end{figure}

\begin{figure}[h]
    \centering
    \includegraphics[width=\linewidth]{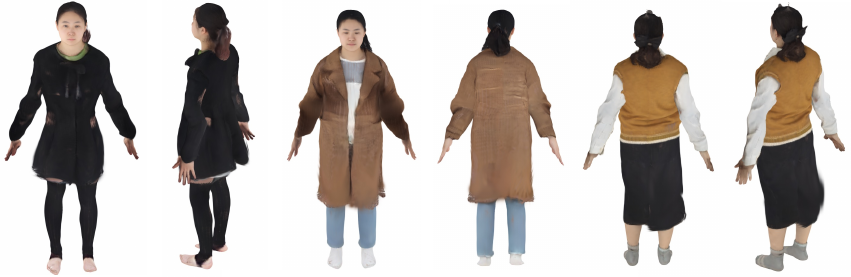}
    \caption{Dress and skirt generation results.}
    \vspace{-3mm}
    \label{fig:dress}
\end{figure}

\noindent\textbf{More Generation Results.}
By training on the composition of three datasets (THuman2.0, THuman2.1, and CustomHuman), our method is capable of generating more diverse avatars and challenging clothing which demonstrates the promise of our methods to achieve better results on larger datasets, as illustrated in Fig.~\ref{fig:extend} and Fig.~\ref{fig:dress}.

\begin{figure}[h]
    \centering
    \includegraphics[width=\linewidth]{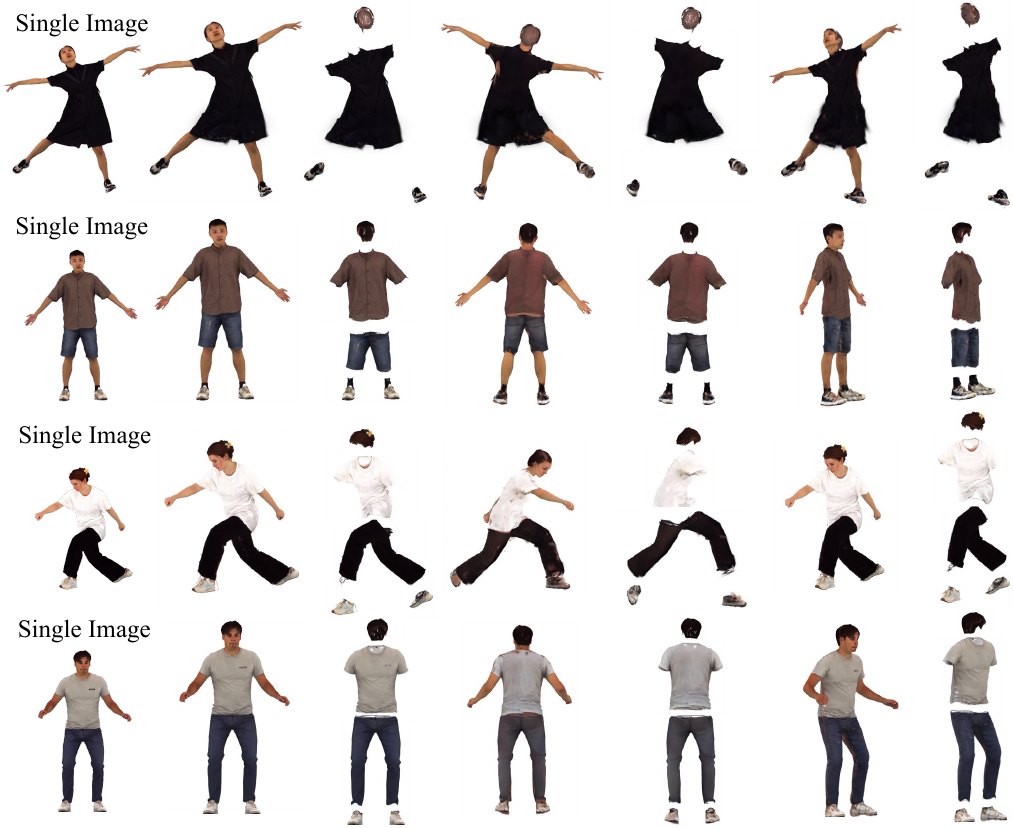}
    \caption{Single image reconstruction on unseen 4DDress dataset.}
    \label{fig:svrecon}
\end{figure}

\noindent\textbf{Generalization capability.}
As shown in Fig~\ref{fig:svrecon}, our method can generalize to various clothing types (dress) and poses (hitting) exhibiting distinctive fingers, facial details, and disentanglement capability. We believe that better results can be obtained when extending to a larger training set.

\noindent\textbf{Ablation on TVLoss.}
Shown in Fig.~\ref{fig:extend}, without TVLoss, the inner body color will be affected by outer components. 

\begin{figure}[h]
    \centering
    \includegraphics[width=\linewidth]{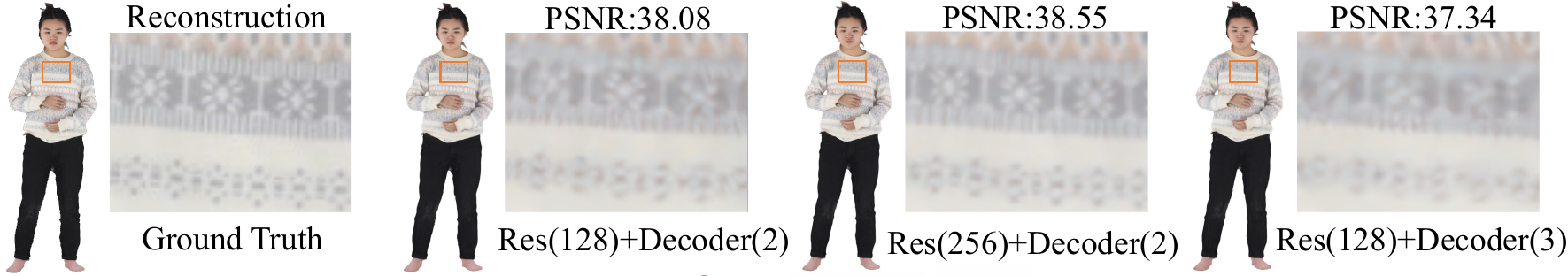}
    \caption{Ablation on feature map resolution and decoder depth.}
    \label{fig:compo}
\end{figure}

\noindent\textbf{Ablation on UV feature plane resolution and decoder depth.}
We conduct a simple experiment to explore the influence of UV feature plane resolution and decoder depth on the results. We reconstruct digital avatars utilizing multi-view images with different resolutions and decoder depths. Demonstrated by Fig.~\ref{fig:compo}, the increase of feature map resolution can enhance the final results by a small margin.  Increasing decoder depth, however, hinders optimization and reduces results. We consider imbalanced data distribution(over 90\% plain texture clothing) and occlusion from limited poses might be the primary constraint of the generation quality for our method. Incorporating video and synthetic data might be helpful to construct a more powerful model.

\end{document}